\documentclass[letterpaper]{article} 
\usepackage{aaai2026}  
\usepackage{times}  
\usepackage{helvet}  
\usepackage{courier}  
\usepackage[hyphens]{url}  
\usepackage{graphicx} 
\usepackage{natbib}  
\usepackage{caption} 
\usepackage{algorithm}
\usepackage{algorithmic}

\usepackage{booktabs} 
\usepackage{tabularx} 
\usepackage{longtable} 
\usepackage{multirow} 

\usepackage{newfloat}
\usepackage{listings}
\DeclareCaptionStyle{ruled}{labelfont=normalfont,labelsep=colon,strut=off} 
\floatstyle{ruled}
\newfloat{listing}{tb}{lst}{}
\floatname{listing}{Listing}
\title{Overcoming the ‘Impracticality’ of RAG: Proposing a Real-World Benchmark and Multi-Dimensional Diagnostic Framework}

\author{
    Kenichirou Narita\textsuperscript{\rm 1}, 
    Siqi Peng\textsuperscript{\rm 1},
    Taku Fukui\textsuperscript{\rm 1},
    Moyuru Yamada\textsuperscript{\rm 1},
    Satoshi Munakata\textsuperscript{\rm 1},
    Satoru Takahashi\textsuperscript{\rm 1}
}
\affiliations{
    \textsuperscript{\rm 1}Artificial Intelligence Laboratories, Fujitsu Limited\\

    k.narita@fujitsu.com,\\
    peng.siqi@fujitsu.com,\\
    fukui.taku@fujitsu.com,\\
    yamada.moyuru@fujitsu.com,\\
    munakata.satoshi@fujitsu.com,\\
    t.satoru@fujitsu.com
}

\usepackage{bibentry}

\begin{document}

\maketitle

\begin{abstract}


Performance evaluation of Retrieval-Augmented Generation (RAG) systems within enterprise environments is governed by multi-dimensional and composite factors extending far beyond simple final accuracy checks. These factors include reasoning complexity, retrieval difficulty, the diverse structure of documents, and stringent requirements for operational explainability. Existing academic benchmarks fail to systematically diagnose these interlocking challenges, resulting in a critical gap where models achieving high performance scores fail to meet the expected reliability in practical deployment. 

To bridge this discrepancy, this research proposes a multi-dimensional diagnostic framework by defining a four-axis difficulty taxonomy and integrating it into an enterprise RAG benchmark to diagnose potential system weaknesses. 
\end{abstract}
\section{Introduction}
Retrieval-Augmented Generation (RAG) \cite{rag} has emerged as the de facto paradigm for overcoming the inherent limitations of Large Language Models (LLMs) in factual reliability and knowledge cutoff. This architectural shift, which grounds LLMs in external documents, is particularly critical in the enterprise. It is fueling the adoption of question-answering systems capable of reasoning over vast proprietary knowledge bases.

While the rapid advancement in RAG research has led to the development of academic benchmarks \cite{visdom,m3docrag}, they fail to capture the practical and multifaceted requirements of enterprise environments. This creates a critical gap, leaving organizations unable to accurately predict a system's real-world performance within their specific context. Consequently, a high score on a conventional benchmark can be dangerously misleading. It often triggers a cycle of costly, ad-hoc modifications, ultimately leading to project delays, service suspensions, and a severe erosion of customer trust at substantial financial cost.

Our analysis pinpoints that this gap stems from three fundamental shortcomings in existing benchmarks, which lack the capacity to evaluate RAG systems on dimensions crucial for enterprise success:

\begin{figure*}
    \centering
    \includegraphics[width=\textwidth]{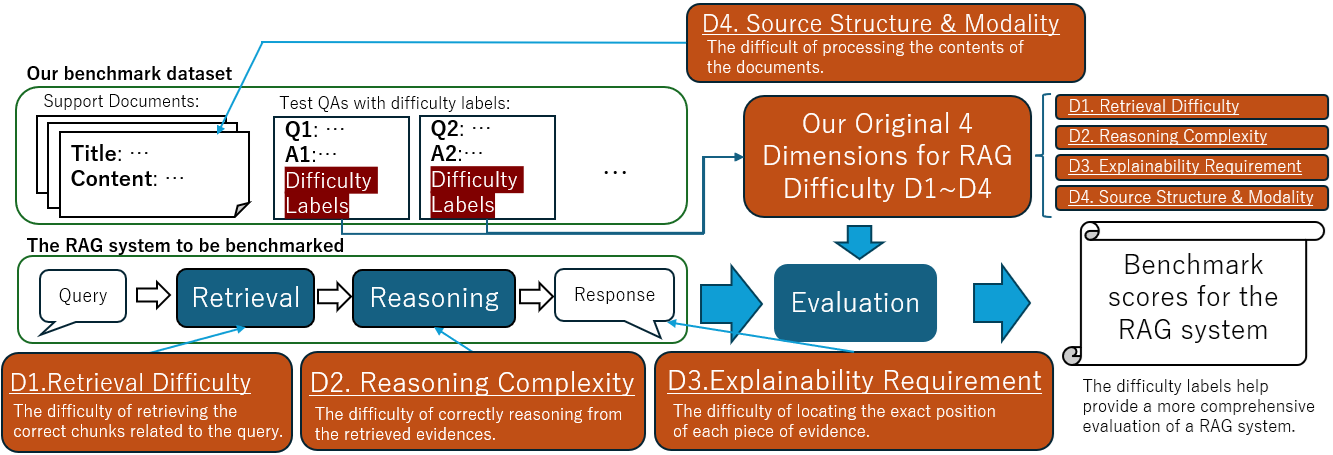}
    \caption{An overview of evaluation framework of our benchmark dataset. Each test query is tagged with difficulty labels based on our original taxonomy of RAG difficulty, which help provide a more comprehensive evaluation of the RAG system.}
    \label{fig:flow}
\end{figure*}

\begin{itemize}
    \item \textbf{Lack of Diagnostic Capability}: Most existing benchmarks treat RAG systems as monolithic black boxes during evaluation, reporting only end-to-end QA accuracy as a measure of overall performance. This approach obscures the ability to systematically diagnose errors, making it difficult to discern whether a failure originates from the retriever's inability to fetch relevant context, or from the LLM's reasoning failure in generating a correct answer.
    \item \textbf{Lack of Evaluation of Composite Tasks}: Existing benchmarks often fail to give a comprehensive evaluation of RAG systems' capability of handling tasks requiring integration of multiple abilities, such as the ability to comprehending complex layouts, tables and charts, and collecting information scattered across a document.
    \item \textbf{Lack of Evidence Traceability}: Most existing benchmarks focus solely on evaluating the correctness of the final output, overlooking critical business requirements for the traceability and reliability of supporting evidence. Specifically, they lack evaluation dimensions that track the sources from which a RAG system generates its output and assess how faithfully the answer is grounded in those evidences.
\end{itemize}

To bridge this gap between academic evaluation and real-world business needs, we propose a novel benchmark dataset grounded in a novel taxonomy of question difficulty. Instead of a single difficulty score, we define and analyze challenges along four distinct axes: \textbf{(i) Reasoning Complexity}, \textbf{(ii) Source Structure \& Modality}, \textbf{(iii) Explainability Requirement}, and \textbf{(iv) Retrieval Difficulty}. This fine-grained metadata allows for precise, practical performance measurement and the diagnosis of bottlenecks for specific question types. An overview of our evaluation framework can be viewed in Figure~\ref{fig:flow}. To demonstrate these capabilities, we also evaluate several leading LLMs and a multi-step reasoning agent on our benchmark.

Our findings reveal that existing LLMs struggle with tasks requiring comprehension of complex business documents and consolidation of multi-source information. While our agent-based approach shows considerable improvements over these baseline LLMs, our benchmark uniquely illuminates more practical failure modes that remain invisible to conventional evaluation methods. This work, therefore, not only presents a new paradigm for RAG evaluation but also identifies critical bottlenecks in the enterprise adoption of LLMs, outlining key directions for future work.

\section{Related Work}
As introduced above, an Enterprise-RAG benchmark must address the unique demands of business environments by capturing real-world complexity, incorporating diagnostic capability and evidence traceability. Existing RAG benchmarks may focus on part of these features, but to the furthest of our knowledge, no previous benchmarks cover all of these. We hereby give a survey on the latest benchmarks which meet part of the demands of business environments. An overview of the features of the these benchmarks can be viewed in Table~\ref{table1}.

\begin{table*}[t]
\centering
\begin{tabularx}{\textwidth}{l|X|l|l|l|X|}
\toprule
\textbf{Dataset} & \textbf{Dimensions of Evaluation} & \textbf{Multi-document} & \textbf{Multimodal} & \textbf{Multi-hop} & \textbf{Evidence Traceability} \\
\midrule
DomainRAG & Generation & true & false & true (48\%) & none \\
ETRQA & Generation & true & false & true (100\%)  & none \\
HeteQA & Generation & true & false & true (18\%)  & text \\
JDocQA & Generation & true & true & true (15\%)  & image \\
Multihop-RAG & Generation \& Retrieval & true & false & true (88\%)   & text \\
VisDoMBench & Generation \& Retrieval & true & true & true (100\%) & text \& image \\
\bottomrule
\end{tabularx}
\caption{An overview of the features of the latest RAG benchmarks.}
\label{table1}
\end{table*}

\subsection{Diagnostic Capability}

The diagnostic capability can be achieved by introducing extra evaluation dimensions. However, most existing benchmarks we surveyed, including HeteQA, ETRQA and DomainRAG, evaluate the RAG systems in one single dimension -- \textit{the generation accuracy}, \textit{a.k.a.}, \textit{the end-to-end QA accuracy}. The performance is measured by the similarity of the gold answer with the response generated by the RAG system. The similarity can be computed via lexical overlap metrics (\textit{e.g.}, ROUGE-L, F1)~\cite{domainrag}, human annotation~\cite{jdocqa}, or evaluated by LLMs (LLM-as-a-judge)~\cite{llmasajudge}.

Besides the most-preferred generation accuracy, Multihop-RAG and VisDoMBench also introduce another dimension -- the \textit{retrieval accuracy}, which reflects how accurately the retrieved chunks cover the ground-truth evidence for the correct answers. The retrieval accuracy can be evaluated using metrics for chunk-level accuracy like \textit{mean reciprocal rank} (MRR)~\cite{multihoprag} or word-level accuracy like \textit{averaged normal longest common sequence} (ANLCS)~\cite{visdom}.

\subsection{Real-world Complexity}

The real-world complexity can be captured by introducing queries requiring complex retrieval tasks. We have summarized three tasks -- \textit{multi-document}, \textit{multi-hop}, and \textit{multimodal} -- that are mainly featured in recent benchmarks. 

\begin{itemize}
\item{\textbf{Multi-document}}{ refers to the task that to retrieve information from multiple \textit{noisy documents}, which are defined to be those relevant to the query but do not contain any information on the answer. With the first benchmarks featuring multi-document retrieval like RECALL~\cite{recall} and RGB~\cite{rgb} being proposed, all latest benchmarks we surveyed have featured multi-document retrieval.}
\item{\textbf{Multi-hop}}{ refers to the task to answer a query that needs reasoning from multiple pieces of evidences\footnote{Note that multi-hop does not necessarily imply multi-document for that the evidences can be from a single document.}. With the first benchmarks featuring multi-hop queries such as Multihop-RAG~\cite{multihoprag} being proposed, all latest benchmarks we surveyed have included multi-hop queries, while the ratio of these queries varies. For example, in HeteQA~\cite{heteqa}, 18\% of all queries are multi-hop, while in ETRQA~\cite{etrqa} and DomainRAG~\cite{domainrag}, the ratio reaches 100\%. }
\item{\textbf{Multimodal}}{ refers to the task to retrieve information from contents incorporating images, videos and/or other non-text formats~\cite{multimodalrag}. Latest multimodal RAG benchmarks include VisDoMBench~\cite{visdom}, JDocQA~\cite{jdocqa}.}
\end{itemize}

While recent benchmarks have increasingly featuring complex retrieval tasks to incorporate real-world complexity, none of them offer a holistic framework for evaluating RAG systems' capability in handling \textit{composite tasks} -- queries that require the integration of multiple retrieval capabilities, such as both multimodal and multi-hop retrieval.

\subsection{Evidence Traceability}

Evidence traceability is achieved by specifying the exact and detailed location of the supporting content that lead to the answer for each query. For text-based references, this can be done through the provenance annotations that pinpoint the location of the evidence -- for example, \textit{``the third paragraph of Chunk 1, starting from...''}. For image-based contents, traceability can be established by providing bounding-box coordinates that precisely frame the relevant tables or figures. 

Among the latest benchmarks we surveyed, Multihop-RAG, HeteRAG and VisDoMBench incorporate evidence traceability for text-based references. VisDomBench and JDocQA support traceability for image-based references.

\section{Problems and Challenges}

As demonstrated in Related Work, latest RAG benchmarks still fail to fully capture the following ``practical challenges'' of Enterprise RAG systems:

\begin{itemize}

\item \textbf{Lack of Diagnostic Capability:} Most existing benchmarks evaluate the RAG systems from a single dimension -- typically end-to-end QA performance -- yielding only one metric, such as Accuracy or F1 Score. This hinders systematic diagnosis of errors, making it difficult to distinguish whether the source of retrieval failure lies in the retriever component or the reasoning capabilities of the LLM.  


\item \textbf{Lack of Evaluation of Composite Tasks:} Although existing benchmarks have increasingly feature complex retrieval tasks, they lack a holistic evaluation framework that evaluates RAG systems' ability in handling complex tasks -- the queries requiring integration of multiple capabilities including multi-step reasoning and complex-layout document comprehension and multi-chunk referencing.


\item \textbf{Lack of Evidence Traceability:} Most benchmarks focus solely on the correctness of the final response of the RAG system, overlooking critical business requirements for the reliability and traceability of supporting evidences. Specifically, they lack specification of the exact location of the supporting contents of each query -- for example, the bounding-box coordinates that precisely identify the figures or tables contributing to the answer.

\end{itemize}

This ultimately leads to the gap between the academic benchmarks and the business ones. To bridge this, we propose a novel evaluation framework that systematically organizes these composite challenges and can identify potential weaknesses in Enterprise-RAG systems -- weaknesses undetectable by existing benchmarks.

\subsection{Taxonomy of Enterprise RAG Difficulty}
We systematically decompose the factors preventing Enterprise RAG systems from answering queries, focusing on the basic RAG architecture (Search and Inference/Generation) and enterprise-specific requirements. Enterprise documents (financial reports, technical manuals, etc.) are complex and often require referencing and parsing figures, complex layouts, and detailed specification tables. Furthermore, enterprise environments frequently require answer evidence for validation. Based on this, we categorize the Enterprise RAG evaluation axes into four dimensions (Table ~\ref{table3-1}).

\begin{table*}[t]
\centering
\begin{tabularx}{\textwidth}{p{2.5cm}|p{2.5cm}|p{3.8cm}|p{3.8cm}|X}
\hline
Dimension & Factors & Elements & Data Format & Evaluation Content \\
\hline
\multirow{3}{*}{\parbox{2.5cm}{Reasoning Complexity}} 
  & Reasoning Depth & Multi-step Reasoning & bool & The amount of LLM logical steps \\
  \cline{2-5}
  & Reasoning Operation & Quantitative Operation\newline Negation Question\newline Cause-and-Effect\newline Comparison (and Conditional Judgment)
  \newline Temporal & bool & Quality of LLM's logical operation \\
  \cline{2-5}
  & Type of Output Processing & - & Exclusive Selection\newline (Summarization,\newline Translation,\newline List Generation) & LLM's answer generation capability \\
\hline
\multirow{3}{*}{\parbox{2.5cm}{Retrieval Difficulty}} 
  & Evidence Location & Multi-document\newline Multi-chunk\newline Low Locality\newline Remote Reference & bool & Retriever's handling of information dispersion \\
  \cline{2-5}
  & Data Scale & Document Volume\newline Chunk Size & bool & Retriever's scalability\\
  \cline{2-5}
  & Query-Evidence Relationship & Abstraction Discrepancy\newline Vocabulary Mismatch & bool & Retriever's semantic search capability \\
\hline
\multirow{2}{*}{\parbox{2.5cm}{Source Structure \& Modality}} 
  & Evidence location & Any Kind of non-plain-text Element\newline Tables/Charts\newline Specific Area of an Element (E.g., a cell of a table or a section of a chart) & bool & LLM's capability of dealing with non-plain-text elements \\
  \cline{2-5}
  & Source document containing complex layout & Nested Logic\newline Long Enumeration List\newline Redundancy & bool & LLM's capability of processing advanced typesetting or structured text (e.g., multi-column formatting and long enumeration) \\
\hline
Explainability Requirement
  & Strictness of Evidence Presentation & - & Exclusive Selection\newline (Coordinate presentation,\newline Multiple locations,\newline Reference hierarchy,\newline No evidence required) & Reliability of answer\newline (Granularity of Evidence) \\
\hline
\end{tabularx}
\caption{Difficulty Definition for Enterprise RAG}
\label{table3-1}
\end{table*}

\subsubsection{Dimension 1: Reasoning Complexity}


This dimension focuses on the LLM's reasoning and generation capability. We evaluate the quantity of logical steps required (Reasoning Depth), the type of reasoning operation at each step (Reasoning Operation), and the difficulty of answer sentence generation (Type of Output Processing). For example, a question requiring multi-step reasoning and comparison/conditional judgment, or those involving quantitative operations (numerical calculation) or negation, frequently act as LLM bottlenecks.   


\subsubsection{Dimension 2: Retrieval Difficulty}

This dimension evaluates the retriever's performance, specifically the difficulty in finding the correct evidence chunk, across three perspectives. Evaluation includes whether evidence is dispersed across multiple documents or deep within a large document (Evidence Location), the total volume of search targets (Data Scale), and the semantic gap between the query and the evidence (Query-Evidence Relationship). Crucially, Abstraction Discrepancy (e.g., a query asking for an abstract concept like "A company's good performance" while the document only contains concrete numbers) necessitates advanced semantic search capabilities.   


\subsubsection{Dimension 3: Source Structure \& Modality}

This dimension evaluates the LLM’s ability to comprehend and analyze document content. It is assessed by determining whether the reference documents contain \textit{complex layouts}, which include any non-plain-text elements such as figures, advanced typesetting (e.g., multi-column formats, header/footer notes), and structured text (e.g., tables, multi-level enumerations). Queries classified as difficult in this dimension often resemble those with high reasoning complexity, where the RAG system may produce incorrect answers despite retrieving the correct chunks. Introducing this dimension enables a more granular analysis of the specific weaknesses in RAG systems.


\subsubsection{Dimension 4: Explainability Requirement}

This dimension evaluates the strictness of practical compliance requirements regarding the reliability and traceability of the answer's grounding in enterprise operations. We define this difficulty based on the required granularity of the evidence. The most rigorous level requires the answer's evidence to be specified by BBOX coordinates (e.g., for rigorous numerical calculation or specific cell extraction from a spec sheet). 

This metric, which can be influenced by not only the query, but also the application use or business requirements, demonstrates the strictness of compliance requirements imposed on the entire RAG system.


\vspace{\baselineskip} 

The difficulty definition systematized in this section functions as an evaluation framework to extract potential RAG system challenges. By tagging the high-difficulty dataset (described next) with these definitions, we can identify the cause of incorrect answers from benchmark results, enabling the specific bottleneck identification required for practical development and deployment.
\section{Proposed Benchmark Dataset}


This research proposes a high-difficulty benchmark dataset based on the difficulty taxonomy developed in the previous section. Source documents were chosen from real-world enterprise materials: Financial Reports, Technical Specifications/Product Catalogs, and Regulatory/Compliance Documents, which are characterized by complex structure and high demands for answer strictness and auditability(Table \ref{table4-1}).

All 100 generated queries are annotated with metadata based on the proposed difficulty definitions, enabling detailed diagnostic analysis of potential RAG system weaknesses undetectable by conventional benchmarks.

\begin{table}[t]
\centering
\begin{tabularx}{\linewidth}{p{5cm}|X}
\toprule
\textbf{Statistic} & \textbf{Value} \\
\midrule
Total Documents & 34 \\
\bottomrule
Total Pages & 1699 \\
\bottomrule
Total Queries & 100 \\
\bottomrule
Chunk Unit & page  \\
\bottomrule
Average Chunks per Doc & 51.485\\
\bottomrule
Avg. Evidence Sources per Query & 2\\
\bottomrule
\end{tabularx}
\caption{Proposed Dataset Information}
\label{table4-1}
\end{table}



\begin{figure}[t]
\centering
\includegraphics[width=0.5\textwidth]{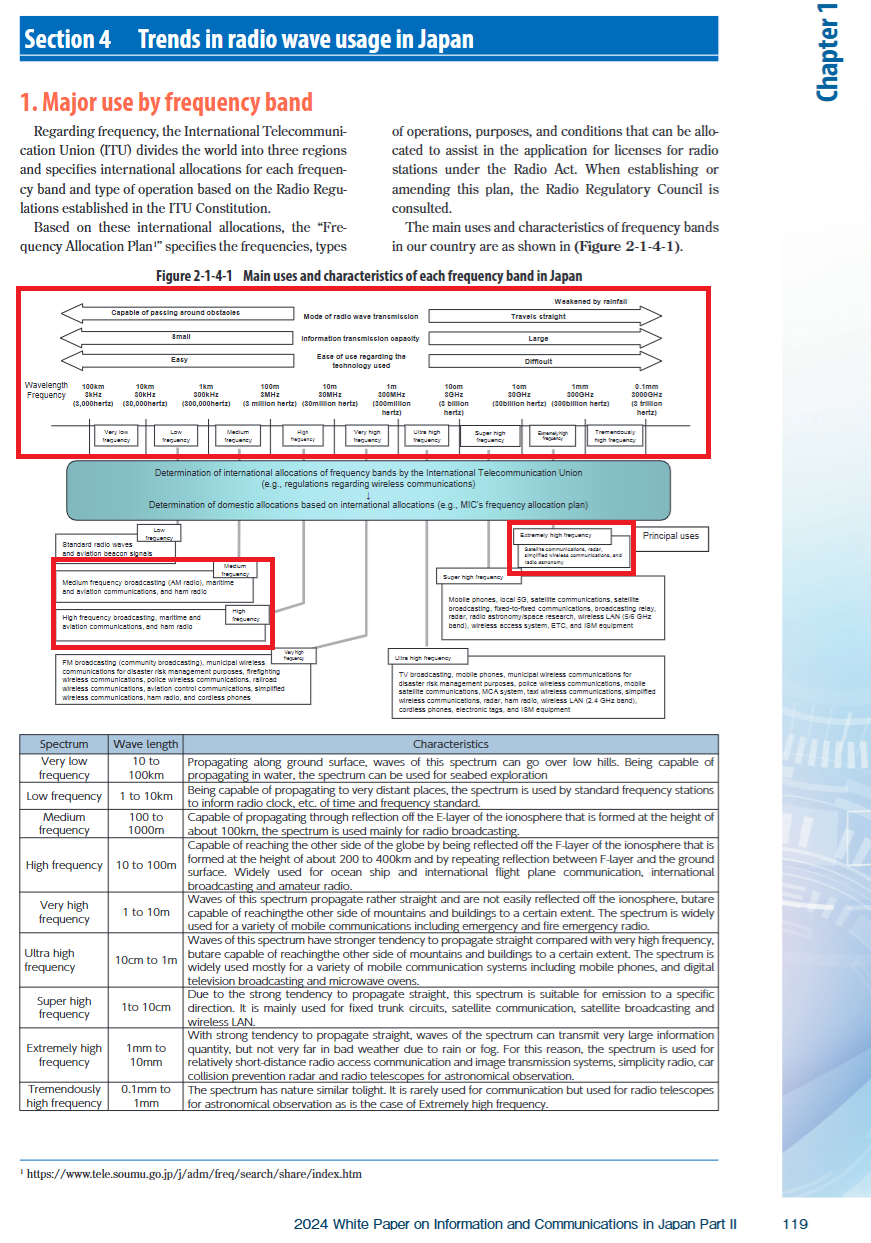} 
\caption{Examples of Evidence Chunk. (Source: Ministry of Internal Affairs and Communications (MIC), Japan. (2024). Information and Communications in Japan White Paper 2024 (English Version). p.133. (\url{https://www.soumu.go.jp/johotsusintokei/whitepaper/eng/WP2024/2024-index.html})).}
\label{fig_no13_evidence_chunk_bbox}
\end{figure}

Query examples are shown in Table \ref{query_example} and Table \ref{query_example_added}. Metadata analysis assigned according to the difficulty definition confirms that these query examples require multi-hop reasoning, comparison, and conditional judgment for the answer. The retrieval is clearly not multi-chunk, suggesting a low difficulty in the retrieval phase. However, as shown in Figure \ref{fig_no13_evidence_chunk_bbox}, the layout of the referenced figure is complex, and the required answer area is highly localized, indicating that the evidence structure is complex. Consequently, decoding the retrieved chunk itself is difficult.

This diagnostic profile proves that the question represents a common composite challenge in practice: retrieval is easy, but structural analysis, multi-step reasoning, and auditability requirements are simultaneously high. This complexity cannot be isolated or diagnosed by conventional single-accuracy evaluations.

\begin{table}[t]
\centering
\begin{tabularx}{\linewidth}{p{1.5cm}|X}
\toprule
\textbf{item} & \textbf{value} \\
\midrule
query & How do airborne communications and satellite communications differ in terms of A. how radio waves propagate, B. the amount of information that can be transmitted, and C. the difficulty of the technologies used? \\
\bottomrule
answer & A. In aviation communications, radio waves can bend around obstacles, while satellite communications exhibit strong line-of-sight characteristics.
B. Satellite communications can transmit larger amounts of information.
C. Satellite communications involve more complex technology. \\
\bottomrule
Bounding Box & top: 33.09, left: 48.84,width: 34.05,height: 25.95...\\
\bottomrule
chunk infomation & Information and Communica-
tions in Japan White Paper 2024 (English Version) :p.133 \\
\bottomrule
\end{tabularx}
\caption{Example questions for the proposed dataset}
\label{query_example}
\end{table}

\begin{table}[t]
\centering
\begin{tabularx}{\linewidth}{p{5cm}|X}
\toprule
\textbf{item} & \textbf{value} \\
\midrule
Reasoning Complexity & \\
\hspace*{0.5cm}Reasoning Depth & True(Multi-step Reasoning)\\
\hspace*{0.5cm}Quantitative Operation & False\\
\hspace*{0.5cm}Negation Question & False\\
\hspace*{0.5cm}Cause and Effect & False\\
\hspace*{0.5cm}Comparison (and Conditional Judgment) & True\\
\hspace*{0.5cm}Temporal Specification & False\\
\hspace*{0.5cm}Type of Output Processing & Summarization\\
\bottomrule
Retrieval Difficulty & \\
\hspace*{0.5cm}multi-document & False\\
\hspace*{0.5cm}multi-chunk & False\\
\hspace*{0.5cm}Low Locality & False\\
\hspace*{0.5cm}Remote Reference & False\\
\hspace*{0.5cm}Document Volume & True(Over 1000 pages)\\
\hspace*{0.5cm}Chunk Size & False(512Token or less)\\
\hspace*{0.5cm}Abstraction Discrepancy & False\\
\hspace*{0.5cm}Vocabulary Mismatch & False\\
\bottomrule
Source Structure \& Modality & \\
\hspace*{0.5cm}Tables/Charts & True\\
\hspace*{0.5cm}Any Kind of Non-plain-text Elements & True\\
\hspace*{0.5cm}Specific Area Reference & True\\
\hspace*{0.5cm}Logical Nesting & False\\
\hspace*{0.5cm}Large-Scale Enumeration & False\\
\hspace*{0.5cm}Redundancy & False\\
\bottomrule
Explainability Requirement & \\
\hspace*{0.5cm}Strictness of Evidence Presentation & Coordinate presentation\\
\bottomrule
\end{tabularx}
\caption{Difficulty Metadata Example}
\label{query_example_added}
\end{table}

\section{Experiment}
The experiment evaluates the effectiveness of the proposed dataset by confirming its high difficulty and verifying its diagnostic capacity to analyze RAG weaknesses using the difficulty metadata.   

The diagnostic evaluation compares two RAG systems against the high-difficulty benchmark :   

\begin{enumerate}
    \item \textbf{MRI-RAG}: Multi-representation-indexing (MRI) is a technique that enhances retrieval robustness by indexing multiple different representations (e.g., summaries, question forms) of the original document chunks. We adopted this technique and implemented the MRI-RAG system based on publicly available LangChain code, which served as the comparative baseline system. Queries are passed unprocessed to the retriever, and the resulting chunks are provided to the LLM for answer generation (zero-shot inference).
    
    \item \textbf{Agentic AI}: A RAG system developed by the authors featuring a multi-step reasoning mechanism composed of three agents :   
    \begin{enumerate}
    \item Query Generation Agent: Analyzes the input query and performs query transformation or decomposition into sub-queries to enhance retriever accuracy.
    \item Filtering Agent: Integrates chunks retrieved via hybrid search (vector and keyword search) using Reciprocal Rank Fusion (RRF), and subsequently determines the utility of each chunk for answering the query. RRF integration serves to ensure robust recall performance and enhance search comprehensiveness by effectively combining multiple distinct search result lists (e.g., results for each sub-query) and fusing the rank of each chunk through scoring.
    \item Answer Agent: Generates the final response using only the chunks deemed useful by the Filtering Agent. 
    \end{enumerate}
\end{enumerate}


The RAG inference component in both systems utilized GPT-4o-mini. For retrieval, we used the text-embedding-3-large model for embedding, setting the retrieval parameter Top-k=10. 

The evaluation used the proposed dataset and JDocQA, a Japanese document QA benchmark with similar complex public documents and task diversity. JDocQA was chosen because it is a Japanese document QA dataset including figures and tables, utilizing data documents such as slides and reports published by Japanese government agencies, and offering a wide range of tasks from Yes/No format to free-form answers, thereby sharing characteristics close to those of the proposed dataset.

\subsection{Diagnostic Metrics}
We employ the conventional Overall Accuracy and the novel Dimensional Diagnostic Accuracy (D-value) based on our difficulty framework :   

\begin{enumerate}
    \item \textbf{Overall Accuracy}: The total correct answer rate, showing general system performance.
    \item \textbf{Diagnostic Accuracy (D-value)}: The correct answer rate restricted only to the subset of questions tagged with "High Difficulty" within a specific dimension. A question is defined as High Difficulty if the sum of its High tags across subcategories exceeds 50\% of the dimension’s total tags ($N$). This D-value quantitatively indicates the degree to which a specific difficulty factor acts as a system bottleneck.
\end{enumerate}


In this experiment, a question group is defined as a High difficulty question group for that dimension if the total number of elements whose values are True (High tags) exceeds a majority (more than 50\%) of the total number of elements, $N$, composing the dimension. This is a provisional arrangement for the current experiment, and the actual determination of "High difficulty" would be configured on a case-by-case basis depending on business needs and client requirements.
\section{Results}

\subsection{Benchmark Difficulty Validation}
\textbf{Table \ref{tab:comparison_accuracy}} compares the Overall Accuracy against JDocQA to confirm the practical difficulty of the proposed dataset.

\begin{table}[htbp]
    \centering
    \begin{tabular}{lcc}
        \toprule
    RAG System & JDoCQA & proposed dataset \\
        \midrule
        MRI-RAG & 0.074 & 0.09 \\
        Agentic AI & 0.149 & 0.170 \\
        \bottomrule
    \end{tabular}
    \caption{Overall Accracy Comparison}
    \label{tab:comparison_accuracy}
\end{table}



MRI-RAG's performance (9.0\%) is comparable to JDocQA (7.4\%), confirming that this dataset imposes a significantly high difficulty level on the baseline model. Agentic AI showed marginally higher Accuracy (17.0\% vs. 14.9\%), suggesting that this dataset contains many questions where multi-step reasoning is beneficial. Both systems remained below 20\% Overall Accuracy, confirming that conventional RAG evaluation only verifies the high difficulty of this benchmark. 

Based on these results, we conclude that this benchmark possesses a high difficulty level that distinguishes it from conventional benchmarks, reflecting the practical challenges of enterprise environments.  

\subsection{Diagnostic Simulation}
The Overall Accuracy stagnating at 17.0\% provides insufficient evidence for identifying the specific strengths and weaknesses behind Agentic AI's limited performance gain, rendering the overall score limited for making practical deployment decisions.

Therefore, in this section, we conduct a diagnostic simulation to identify Agentic AI's strengths and the weaknesses (bottlenecks) that should be the focus of future R\&D, which were hidden behind this limited performance difference.

The Diagnostic Accuracy (D-value) based on the proposed difficulty definitions is shown in Table \ref{tab:d_accuracy}.

\begin{table}[htbp]
    \begin{tabularx}{\linewidth}{p{3cm}|p{2cm}|X}
        \toprule
        \textbf{Dimension} & \textbf{MRI-RAG} & \textbf{Agentic AI} \\
        \midrule
        Reasoning Complexity & 2.5\% & 5.0\%(+2.5pt)\\
        Retrieval Difficulty & 0.0\% & 12.0\%(+12.0pt) \\
        Source Structure \& Modality & 10.0\% & 20.0\%(+10.0pt) \\
        Explainability Requirement & 9.1\% & 17.2\%(+8.1pt) \\
        \bottomrule
    \end{tabularx}
    \caption{Dimensional Diagnostic Metrics (D-value) Comparison}
    \label{tab:d_accuracy}
\end{table}

The analysis of the D-value quantitatively demonstrated the primary effects of Agentic AI. Agentic AI showed a dramatic 12.0 point improvement in Retrieval Difficulty, increasing from a baseline of 0.0\% to 12.0\%. This result quantitatively demonstrates that the primary effect of Agentic AI's multi-step reasoning mechanism is concentrated not on the final inference by the LLM, but on the optimization of search queries through query analysis and the robust processing of multi-chunk search results via RRF integration. The agent mechanism improves system performance by assisting the retrieval phase to ensure the LLM receives more appropriate context. Furthermore, improvement was also seen in Source Structure \& Modality, increasing 10.0 points from 10.0\% to 20.0\%. This improvement is presumed to be attributed to the fact that the retrieval stage accurately filtered more structurally useful chunks and provided them to the Answer Agent, thereby improving the accuracy of structural analysis on the retrieved evidence.

Conversely, the introduction effect of Agentic AI was most limited in Reasoning Complexity. This D-value remained at only 5.0\% for Agentic AI (an improvement of +2.5pt). This result clearly shows that the LLM's final information integration and reasoning capacity remains the greatest constraint (bottleneck) of the system in question groups where reasoning steps are complicated, or quantitative calculation or negation judgment is required. Despite Agentic AI substantially mitigating retrieval phase challenges, the overall performance stagnation is due to this constraint in reasoning capacity.

Thus, by utilizing the Dimensional Diagnostic Metrics, it was made clear that Agentic AI's retrieval performance improved, while its reasoning capacity did not increase significantly. This enabled the explicit identification of improvements and problems within the RAG system, which was impossible with conventional Overall Accuracy. This diagnostic result suggests the necessity for future system improvement in Agentic AI to shift its focus towards the "development of a mechanism specialized in answering complex, enterprise-specific queries by concentrating on query complexity."
\section{Limitations}
To ensure academic rigor, we acknowledge the following limitations :

\begin{enumerate}
\item \textbf{Constraints on Dataset Scale and Evaluation Scope:} The current query count (100 questions) is a first step toward comprehensive evaluation but is limited compared to large-scale academic benchmarks. This scale restricts the assessment of question distribution in specific domains or document structures, thus limiting the evaluation of generalization capability. Future work must involve the progressive expansion of the dataset with diverse enterprise documents based on the proposed difficulty taxonomy to enhance statistical significance.

\item \textbf{Unassessed Trade-offs in Cost and Efficiency:} Evaluation lacks axes for practical efficiency (latency, token usage). Given that Agentic AI performs multi-step reasoning, quantifying the trade-off between diagnostic performance and operational cost is required for realistic implementation planning.

\item \textbf{Unevaluated Interacting Effects among Difficulty Dimensions:} Although the four dimensions are defined to be orthogonal, the dynamic interaction and synergistic failure modes resulting from multiple concurrent high-difficulty tags require further rigorous analysis and modeling.

\item \textbf{Dependency on Specific LLM Foundation Models:} Experimental results are dependent on the chosen LLM (GPT-4o-mini) and retriever architecture. To enhance the benchmark’s generalizability and prevent bias in difficulty definition, comprehensive validation using a diverse range of LLM foundations is required.
\end{enumerate}
\section{Conclusion}
This research aims to diagnostically identify the performance bottlenecks in RAG systems facing composite challenges in real-world enterprise environments. 

We proposed a multi-dimensional evaluation framework that systematically defines question difficulty across four independent dimensions: Reasoning Complexity, Retrieval Difficulty, Source Structure \& Modality, and Explainability Requirement. By combining this framework with a high-difficulty benchmark dataset sourced from financial reports and technical specifications, a new evaluation system capable of diagnosing the true capacity of RAG systems was established.

Based on this new evaluation system, we diagnosed the Agentic AI developed by the authors and succeeded in separating the performance bottlenecks that were indistinguishable by conventional single metrics. Specifically, Agentic AI's agent mechanism demonstrated a strong focus on significant improvement in retrieval performance, while quantitatively showing that the LLM's reasoning capacity remains the greatest constraint on the overall system. 

This diagnostic finding demonstrates the necessity of focusing future system improvement efforts on enhancing reasoning ability tailored to query complexity. The proposed framework presents a new, practical paradigm for RAG evaluation that enables the proactive identification of potential performance issues and allows for the determination of development guidelines and strategic efficiency improvement.

The benchmark dataset is slated for public release on GitHub. Future work will focus on modeling the dynamic interactions between difficulty dimensions and incorporating composite evaluations that include latency and cost factors, ultimately evolving the framework into a comprehensive set of RAG system selection guidelines for the enterprise sector.

\section{Acknowledgments}
The authors sincerely thank all individuals and institutions who provided invaluable support and cooperation in conducting this research and establishing this new diagnostic framework for Enterprise RAG evaluation.

\bigskip

\bibliography{aaai2026}


\end{document}